\documentclass{article}
\usepackage[table,xcdraw,dvipsnames]{xcolor}

\usepackage[preprint]{corl_2026} % Use this for the initial submission.
\usepackage{outlines}

\usepackage{enumitem}
\setenumerate[2]{label=\alph*.}
\setenumerate[3]{label=\roman*.}
\setlist[enumerate]{itemsep=2pt, parsep=0pt, topsep=2pt, partopsep=0pt}
\setlist[itemize]{itemsep=2pt, parsep=0pt, topsep=2pt, partopsep=0pt}
\usepackage{microtype}
\usepackage{graphicx}
\usepackage{caption}
\usepackage{subcaption}
\usepackage{booktabs}
\usepackage{hyperref}
\usepackage{amsmath}
\usepackage{amssymb}
\usepackage{mathtools}
\usepackage{amsthm}
\usepackage[capitalize  ]{cleveref}
\usepackage{xspace}
\usepackage[normalem]{ulem}
\usepackage{algorithm}
\usepackage{algpseudocode}
\usepackage{wrapfig}

\crefname{equation}{Eq.}{Eqs.}
\crefname{figure}{Fig.}{Figs.}
\crefname{tabular}{Tab.}{Tabs.}
\crefname{section}{Sec.}{Secs.}
\crefname{algocf}{Alg.}{Algs.}
\crefname{algocfline}{Line}{Lines}
\crefname{lem}{Lemma}{Lemmata}
\crefname{appendix}{Appx.}{Appxs.}
\crefname{algorithm}{Alg.}{Algs.}
\theoremstyle{plain}

\theoremstyle{definition}

\theoremstyle{remark}

\newcommand{\mpvimodule}[2]{\textcolor[rgb]{#1}{\textbf{#2}}}
\newcommand{\subtp}[1]{\mpvimodule{0.3,0.7,0.55}{#1}}
\newcommand{\orch}[1]{\mpvimodule{0.3,0.3,0.9}{#1}}
\newcommand{\nav}[1]{\mpvimodule{0.99,0.1,0.1}{#1}}
\newcommand{\manip}[1]{\mpvimodule{0.6,0.3,0.7}{#1}}
\newcommand{\compchk}[1]{\mpvimodule{0.9,0.4,0.0}{#1}}

\newcommand{\ourmethod}{MPVI\xspace}

\newcommand{\instruction}{\ell}
\newcommand{\plan}{\mathcal{P}}
\newcommand{\segment}{s}
\newcommand{\category}{c}
\newcommand{\subtaskdescription}{\delta}
\newcommand{\target}{t}
\newcommand{\completioncriteria}{\phi}

\title{Make Your VLA More Robust Without More Data By Interleaving Motion Planning}

\author{
  Dan B.W. Choe\thanks{Equal contribution. Corresponding authors: \texttt{\{bchoe7}, \texttt{ssangeetha3\}@gatech.edu}.\\ All authors are with the Georgia Institute of Technology, Atlanta, GA, USA},\;\;Sundhar Vinodh Sangeetha\footnotemark[1],\;\;Samuel Coogan,\;\;Shreyas Kousik\\
  Georgia Institute of Technology
}

\begin{document}
\maketitle

%===============================================================================
\begin{abstract}
% VLAs have achieved amazing things (gotta butter up the community to start out), but  they are still bad at long-horizon tasks
Vision-Language-Action (VLA) models have shown remarkable progress for mobile manipulation, but their performance on long-horizon tasks remains poor.
These tasks are especially challenging because (1) progress toward high-level goals must be maintained across extended sequences of spatially distributed subtasks, and (2) early execution errors compound rapidly over the task horizon.
These challenges persist despite finetuning on large human teleoperated mobile manipulation data, indicating that more data alone may not resolve the problem. 
To address these challenges, we propose \ourmethod: Motion Planner~/~VLA Interleaving, a framework that integrates model-based motion planning with VLAs to improve robustness without further training.
The proposed integration enables localization and navigation to distant or occluded target objects through cluttered scenes using open-vocabulary object detection, frontier exploration and motion planning.
However, such integration is non-trivial, requiring reliable switching between modules; we show one way forward via VLM-based completion checking with proprioceptive triggers.
We evaluate our approach on the BEHAVIOR-1K benchmark and demonstrate 113\%
improvement in task progress over a top end-to-end VLA baseline.
Additional details are available at the project page: 
\href{https://mpvi.netlify.app/}{https://mpvi.netlify.app/}.

\end{abstract}


%===============================================================================

\section{Introduction}

Long-horizon mobile manipulation, especially in the context of household robotics tasks, remains a significant challenge for several reasons: household scenes are cluttered, contain diverse objects and long-horizon tasks require robots to execute extended sequences of spatially distributed subtasks while maintaining progress towards a high level goal.
A current paradigm for addressing these challenges is to train large Vision-Language-Action (VLA) models that output low-level robot actions using natural language task descriptions and camera images.
These models rely on internet-scale pretraining and large robotics datasets \cite{bridgedata, OpenX, droid, mees2022calvin} to learn policies that can generalize across tasks and environments.

% As a canonical example, the BEHAVIOR-1K benchmark \cite{BEHAVIOR-1K} provides 50 household tasks that involve long-horizon mobile manipulation, along with 1,200 hours of human teleoperated training data across 10,000 demonstrations of these tasks.
% Despite this large dataset and a variety of training methods developed by participants in the 2025 BEHAVIOR-1K Challenge \cite{behavior-first-place,bai2025comet}, VLAs achieve poor performance in the benchmark, with the best competition solution achieving only 11\% average success rate across all tasks.
As a canonical example, the BEHAVIOR-1K benchmark~\cite{BEHAVIOR-1K} provides 1,000 household tasks that involve long-horizon mobile manipulation.
The 2025 BEHAVIOR Challenge at NeurIPS~\cite{behavior2025challenge} selects a 50-task subset for evaluation and releases 1,200 hours of human teleoperated training data across 10,000 demonstrations of these tasks.
Despite this large dataset and a variety of training methods developed by participants in the 2025 BEHAVIOR Challenge~\cite{behavior-first-place,bai2025comet}, VLAs achieve poor performance in the benchmark, with the best competition solution achieving only 12.4\% average success rate across all tasks.

A recent systematic failure analysis of the top BEHAVIOR-1K solutions uncovers failure modes such as navigation failures, collisions, poor spatial grounding, and execution-order confusion \cite{rasouli2026vlas}.
These failures point to architectural challenges in long-horizon memory, task progress tracking, and spatial reasoning that are difficult to resolve within a purely end-to-end VLA framework.
Many of these failure modes correspond to problems that classical planning methods are explicitly designed to address, such as navigating to distant objects and computing collision-free paths through obstacle-dense scenes.
In this work, we ask how classical motion planning should be integrated with Vision-Language Action models to improve success rates over end-to-end VLAs on long-horizon mobile manipulation tasks, and present a recipe for doing so.

\begin{figure*}[t]
    \centering
    \includegraphics[width=\linewidth]{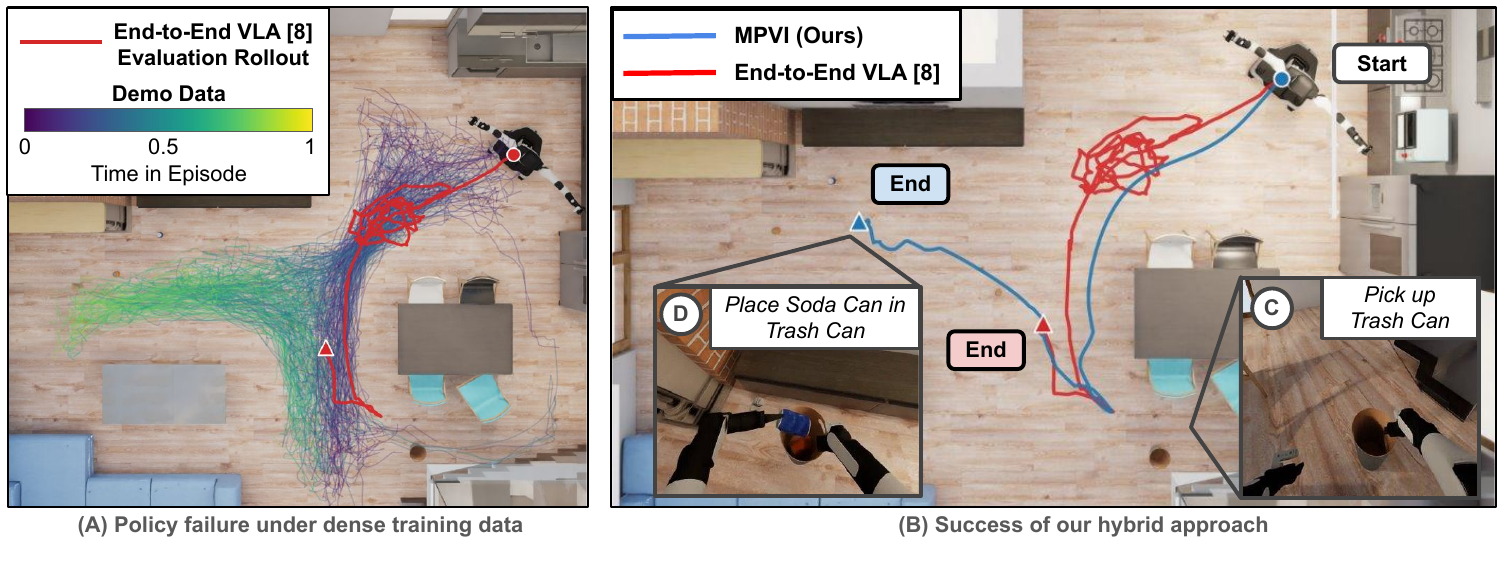}
    \caption{\textbf{(A)} An evaluation trajectory rollout of the baseline end-to-end VLA model (red) is overlaid against the extensive set of expert demonstration training data (color-coded by episodic time). Although the evaluation task lies entirely within the high-density support of the training distribution, the end-to-end VLA exhibits task failure. This highlights a fundamental structural bottleneck that cannot be mitigated simply by scaling training data.
    \textbf{(B)} \ourmethod integrates classical motion planning for navigation with a VLA for manipulation, improving task success in long-horizon mobile manipulation. On a task requiring the robot to locate and grasp a trash can with one hand (C), then collect soda cans from multiple locations and place them into the held trash can (D), an end-to-end VLA baseline (red) is unable to successfully complete the task due to navigation failure and execution order confusion. Our method (blue) applies classical planning for navigation, avoiding such failure modes and enabling reliable task completion.
     }
    \label{fig:front figure}
\end{figure*}

While previous works have integrated classical motion planners with VLAs, such methods have primarily used motion planning for short, pre-grasp motions with fixed-base VLAs on short-horizon pick-and-place tasks \cite{wu2025moto,he2025falconactivelydecoupledvisuomotor,anywherevla}.
These works also motivate the use of classical motion planning by citing a lack of large-scale mobile manipulation datasets.
However, we consider long-horizon, complex mobile manipulation tasks where state-of-the-art whole-body VLA models achieve low performance despite finetuning on the large teleoperated training datasets provided in the BEHAVIOR-1K benchmark.
Together, these results encourage our investigation into hybrid approaches that leverage the complementary strengths of learned policies and classical planners.
In summary, we make the following contributions:

% \sundhar{TODO: 33.32\% of subtasks in human annotated BEHAVIOR-1K training data are navigation \cite{bai2025comet}, this is XX\% of the time duration of the training data. BEHAVIOR-1K tasks are a representative sample of household tasks (human survey). }

% \subsection{Contributions}
\begin{enumerate}
    \item We develop a framework, \ourmethod: Motion Planner / VLA Interleaving, that carefully integrates a motion planner with a VLA. We demonstrate that a motion planner can serve as a steering mechanism for VLAs, mitigating failures in long-horizon mobile manipulation tasks.
    \item We introduce a subtask transition mechanism that grounds semantic visual completion checking in physical proprioception. We identify that relying solely on vision-language reasoning for task transitions leads to cascading failures driven by VLM hallucinations.
    \item We show through a thorough experimental evaluation on the BEHAVIOR-1K benchmark, comprising realistic long-horizon household robotics tasks in a high-fidelity simulator, that \ourmethod improves task success by addressing failure modes of a SoTA end-to-end VLA baseline. 
\end{enumerate}

% The remainder of this paper is organized as follows.
% In \Cref{sec: related work} we review techniques used by state-of-the-art end-to-end VLA approaches to solve long-horizon mobile manipulation tasks as well as recent attempts to integrate classical motion planning with VLAs.
% Then, we present our method for this integration in \Cref{sec: method}.
% In \Cref{sec: experiments}, we perform an experimental evaluation that compares our hybrid approach to end-to-end VLA approaches on the BEHAVIOR-1K dataset of long-horizon household robotics tasks.
\section{Related Work}
\label{sec: related work}

We now discuss works on long-horizon mobile manipulation tasks specified in natural language, focusing on state-of-the-art end-to-end VLA approaches and methods to integrate classical motion planning methods with VLAs.

\subsection{VLAs for Long-Horizon Mobile Manipulation}

VLAs struggle to maintain coherent behavior over extended task sequences due to errors that propagate and compound across subtasks \cite{skilldiscovery, chen2024scar}.
Task decomposition and large-scale data aggregation are the main strategies used to address this issue \cite{fan2025longvlaunleashinglonghorizoncapability,wen2025dexvla,pi0.5}.

\paragraph{Task Decomposition} 
Recent works decompose complex tasks into shorter subtasks using the semantic knowledge of LLMs and VLMs \cite{ahn2022can,hu2023look,li2023interactive,liu2024ok}.
For example, $\pi_{0.7}$, a state-of-the-art VLA, first predicts high-level semantic subtasks using a separate learned policy then generates low-level actions conditioned on this prediction with the VLA model \cite{intelligence2026pi}.
Since open-source code for $\pi_{0.7}$ has not been published at the time of writing, we adopt the $\pi_{0.5}$ VLA \cite{pi0.5} as the manipulation policy in our method. 
Leading BEHAVIOR-1K solutions~\cite{behavior-first-place,bai2025comet} use $\pi_{0.5}$ in an end-to-end manner.
Considering navigation and manipulation separately in task decomposition of mobile manipulation tasks has been shown to yield improvements.
For example, Long-VLA \cite{fan2025longvlaunleashinglonghorizoncapability} segments subtasks into moving and interaction phases and masks input to the VLA dependent on the current phase, allowing it to focus on phase relevant inputs and improving performance.

\paragraph{Scaling Data}
VLA performance on long-horizon tasks also depends strongly on the scale and diversity of training data.
Recent efforts expand robot learning data through multi-robot collection \cite{bridgedata, droid}, cross-embodiment transfer \cite{OpenX}, and internet-scale non-robot data \cite{pi0.5}.
BEHAVIOR-1K competition solutions follow a similar direction, expanding the training dataset by introducing random pose perturbations and retaining successful rollouts from the pre-trained policy, and generating approximately 3,600 additional manipulation trajectories using a motion planner \cite{bai2025comet}.

\subsection{Integrating Motion Planners with VLAs}
Several recent works have explored integrating classical motion planning with VLAs to tackle mobile manipulation tasks.
In general, these works consider integration with fixed-base VLAs that control only the robot upper body, as opposed to whole-body VLAs, motivating their approach by citing a lack of sufficient training data for whole-body mobile manipulation ~\cite{momanipvla,wu2025moto,he2025falconactivelydecoupledvisuomotor,anywherevla}.
In contrast, we show that the integration of classical motion planning provides superior performance on mobile manipulation tasks even with VLAs designed for whole-body control and trained on large-scale whole-body mobile manipulation datasets.

One line of work uses trajectory optimization to coordinate base and arm motion for pre-grasp navigation \cite{momanipvla,wu2025moto}.
These methods extract a target end-effector pose, either from the VLA's action output or by using a Vision-Language Model (VLM) to identify contact points on the robot and target object, and plan a trajectory to reach it.
However, this integration is inherently limited to short-range motion while the robot is already positioned near the target object.
These approaches cannot perform navigation across or even within rooms, restricting their applicability to a narrow segment of mobile manipulation tasks.
% FALCON~\cite{he2025falconactivelydecoupledvisuomotor} trains separate diffusion policies for locomotion and manipulation, coordinated through shared VLM embeddings. While this decoupled architecture aligns with our intuition, FALCON employs learned policies for both subsystems rather than classical motion planning, and evaluates on short-horizon quadruped tasks rather than long-horizon household mobile manipulation.

Closest to our work, AnywhereVLA \cite{anywherevla} combines frontier exploration and motion planning with a fixed-base VLA manipulation policy, navigating to a pose relative to the target object before executing manipulation.
However, this method lacks detection of manipulation subtask completion and does not perform task decomposition, limiting it to short horizon tasks. Its evaluation consists solely of instructions of the form ``pick up the [object] and place it in the [area]'' and ``bring the [object] to [location].''
In contrast, we couple task decomposition with proprioception-triggered completion checking, enabling reliable handoffs between navigation and manipulation across arbitrarily many subtasks and thus scaling to the long-horizon, multi-stage household tasks in BEHAVIOR-1K.

\section{Problem Formulation}
% We consider long-horizon mobile manipulation tasks.
% Each task is specified by a natural language instruction $\instruction$ (e.g., ``put the soda cans in the living room in the trash can in the kitchen'').
% At each timestep $t$, the robot observes its proprioceptive state $q_t$ and RGBD images from head- and wrist-mounted cameras.
% We assume access to a map of the environment \textit{a priori} \shrey{need to justify this with a brief statement and citation to some SLAM stuff}.

% Task completion is defined using the Behavior Domain Definition Language (BDDL), which specifies goal conditions as logical predicates over object states; these BDDL goal conditions are not made available to the robot, and are only used for evaluation.
% \shrey{cite something (behavior1k?) to justify that this problem setup is reasonable/common}
% A task is successful if the robot achieves a configuration satisfying the predicates within a fixed time horizon.
% \shrey{rephrased the previous so a reviewer doesn't misunderstand these as assumptions}
We consider long-horizon mobile manipulation tasks.
Each task is specified by a natural language instruction $\instruction$ (e.g., ``put the soda cans in the living room in the trash can in the kitchen'').
At each timestep $t$, the robot observes its proprioceptive state $q_t$ and RGBD images from head- and wrist-mounted cameras.
We assume access to a map of the environment \textit{a priori}; a reasonable assumption in household settings, where robust semantic mapping and localization methods are well established~\cite{raychaudhuri2025semantic,songsemantic} but do not assume knowledge of task object locations.

Task completion is defined using the Behavior Domain Definition Language (BDDL)~\cite{BEHAVIOR-1K,behavior2021}, which specifies goal conditions as logical predicates over object states; these BDDL goal conditions are not made available to the robot, and are only used for evaluation.
A task is successful if the robot achieves a environment configuration satisfying the predicates within a fixed time horizon.
\section{Method}
\label{sec: method}

\begin{figure*}[ht]
    \centering
    \includegraphics[width=\textwidth]{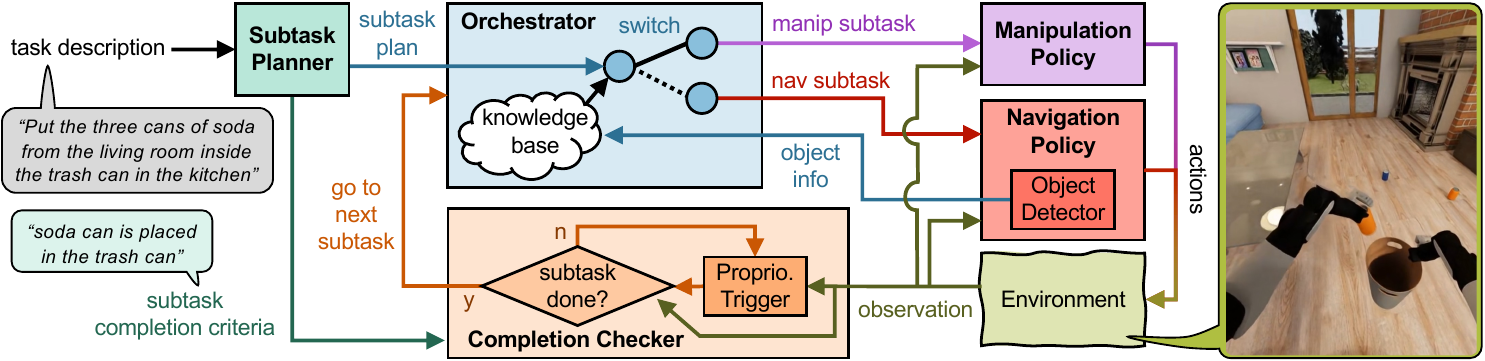}
    \caption{Overview of the proposed method.
    We design an Orchestrator to switch between a Manipulation Policy (VLA) and a Navigation Policy (classical motion planner) depending on the subtask plan generated by a Subtask Planner (LLM).
    The Navigation Policy also returns environment object information to the Orchestrator to keep track of subtask progress for its switching logic.
    For the system to know when to query the Orchestrator, we design a Completion Checker that uses a proprioceptive trigger to decide when to check if a subtask is complete (by querying a VLM).}
    \label{fig: block diagram}
\end{figure*}

\Cref{fig: block diagram} shows the overall \ourmethod framework and its control flow.
It comprises five modules:
(1) a \subtp{Subtask Planner} (LLM), (2) an \orch{Orchestrator},
(3) a \nav{Navigation} module (motion planner), (4) a \manip{Manipulation} module (VLA), and (5) a \compchk{Completion Checker} (VLM).
\ourmethod is deliberately modular, treating each component as a black box and remaining agnostic to the specific LLM, VLM, VLA, or motion planner used.
Our contribution is therefore at the integration level, addressing how a motion planner and a VLA should be composed to solve long-horizon mobile manipulation. The framework benefits from improvements to any of its constituent modules.
The remainder of this section describes the role of each of the five modules, with implementation details deferred to \Cref{app: method details}.

\paragraph*{\subtp{Subtask Planning}}
\label{sec: subtask}
End-to-end VLAs compound errors over long horizons~\cite{skilldiscovery, chen2024scar}.
We prompt the LLM to decompose the task instruction $\instruction$ into short subtasks, each tagged as navigation or manipulation. 
Each subtask also carries a \emph{state-based} completion criterion, a predicate over the world state (e.g., ``the can is in the trash'') rather than an action predicate (e.g., ``the robot placed the can''), because the former is verifiable from a single image, whereas the latter requires temporal reasoning over a sequence of frames. 
Full prompt, fine-tuning, and post-processing details are in \Cref{app: subtask planning details}.

\paragraph*{\orch{Orchestrator}}
\label{sec: integration}
% The orchestrator has two jobs: it dispatches subtasks between the motion planner and VLA, and it tracks which segments have already been completed.
% It walks through the task plan $\plan$ in order, dispatching each navigation subtask to the motion planner on its own and grouping consecutive manipulation subtasks into a single execution unit that is handed to the VLA along with the completion criterion of the last subtask in the group.
The orchestrator serves two functions: (i) routing subtasks to the motion planner or the VLA, and (ii) tracking subtask completion.
It processes the plan $\plan$ sequentially, dispatching each navigation subtask individually and grouping consecutive manipulation subtasks into a single execution unit passed to the VLA together with the completion criterion of the final manipulation subtask in the group.
Throughout execution it maintains a knowledge base of rooms, areas, large fixtures, and objects discovered during navigation, enabling subsequent navigation segments targeting the same object to plan directly rather than repeating the visual search. 
If a module fails to complete its subtask within a fixed time budget, the orchestrator marks the subtask failed and advances to the next segment (\Cref{app: orchestrator details}).

\paragraph*{\nav{Navigation}}
\label{sec: motion planning}
%  \begin{wrapfigure}{r}{0.37\textwidth}   % r=right, l=left, 45% column width
%     \centering
%     \vspace{-2.2em}                       % optional: tighten top spacing
%     \includegraphics[width=0.37\textwidth]{figures/Occluded_Figure_Sized - Google Slides.pdf}
%  \caption{\textbf{Long-horizon navigation} Evaluation rollouts of the baseline end-to-end VLA (red) and MPVI (blue) navigating toward a target object (trash can) that is initially occluded from the robot's starting viewpoint (A). The baseline VLA policy fails to progress, indicating a vulnerability to occluded targets due to limited long-horizon spatial reasoning. Our method uses frontier exploration coupled with open-vocabulary object detection (B) to systematically search the environment, successfully uncovering and navigating to the target.}
%   \label{fig:navigation occlusion}

%     \vspace{-2.2em}                       % optional: tighten bottom spacing
%   \end{wrapfigure}

Recent failure analyses of end-to-end VLAs on long-horizon tasks identify navigation errors, collisions, and execution-order confusion as recurring failure modes~\cite{rasouli2026vlas}, precisely the modes that classical motion planning addresses by construction.
We therefore use A* motion planning\footnote{Consistent with our modular design, the specific choice of A* is incidental. Our contribution is the interleaving with the VLA rather than the planner itself, and we expect the gains reported in \Cref{sec: experiments} to compound with more sophisticated classical planners \cite{karaman2011sampling, dolgov2010path}.} for all base motion, using the VLA only inside the manipulation handoff radius around the target.
% \shrey{I am worried that a reviewer will say A* is too simple.
% We need to make it clear that the point is NOT that we are using A*, but rather that we are interleaving a classical motion planner, and just happen to pick A* because it's the simplest instantiation of this one module in our overall framework.
% Since A* works (cite our experiments), we can be confident that our method can be improved even further by using fancier planners, which is not the focus of this work.}
Classical motion planning generally assumes access to an obstacle/traversability map with labeled rooms and accurate robot localization; this is a reasonable assumption in household settings, where robust semantic mapping and localization methods are by now well established~\cite{raychaudhuri2025semantic, songsemantic}.
Finding the target object itself (e.g., a fridge) is handled at runtime: the orchestrator's knowledge base is queried first; if the object location is unknown, then open-vocabulary detection with GroundingDINO~\cite{liu2024grounding} and frontier-based exploration~\cite{yamauchi1997frontier} are applied to search for the object (\Cref{app: navigation details}).

\paragraph*{\manip{Manipulation}}
\label{sec: manipulation}

% \shrey{At this point we have gone two full pages with no figure or math to break the visual flow.
% It's a stupid thing, but CoRL reviewers often just want to see cool pretty stuff.
% Would be great to add a little manipulation figure.
% Potentially condense these subsections together and move some detail to the appendix to fit more figures.}

Manipulation subtasks are executed by a VLA, which receives the subtask description $\subtaskdescription_i$ and RGB images from the robot's head and wrist cameras as input, and outputs robot actions.
In our experiments we use openpi-comet~\cite{bai2025comet} as the manipulation policy without further training.
Openpi-comet is built on $\pi_{0.5}$~\cite{pi0.5} and fine-tuned on the teleoperation dataset released for the 2025 BEHAVIOR Challenge at NeurIPS~\cite{behavior2025challenge}, with additional motion-planner-synthesized trajectories and successful rollouts retained during RL training.
The orchestrator transfers control to the VLA once the robot is within the handoff distance $d_h$ of the target object, and execution continues until the completion checker signals subtask termination.

% \shrey{Cite which VLA we use (i.e. just say for implementation we use BLAH)}
% The VLA architecture, input modalities, action chunk size, and action execution strategy remain unmodified; base motion is permitted if commanded by the VLA.
% The primary challenge in this module is detecting completion of manipulation subtasks to trigger transitions to the navigation module, which we solve next with a completion checker.

\paragraph*{\compchk{Completion Checker}}
\label{sec: completion checker}

\begin{figure*}[ht]
    \centering
    \includegraphics[width=\textwidth]{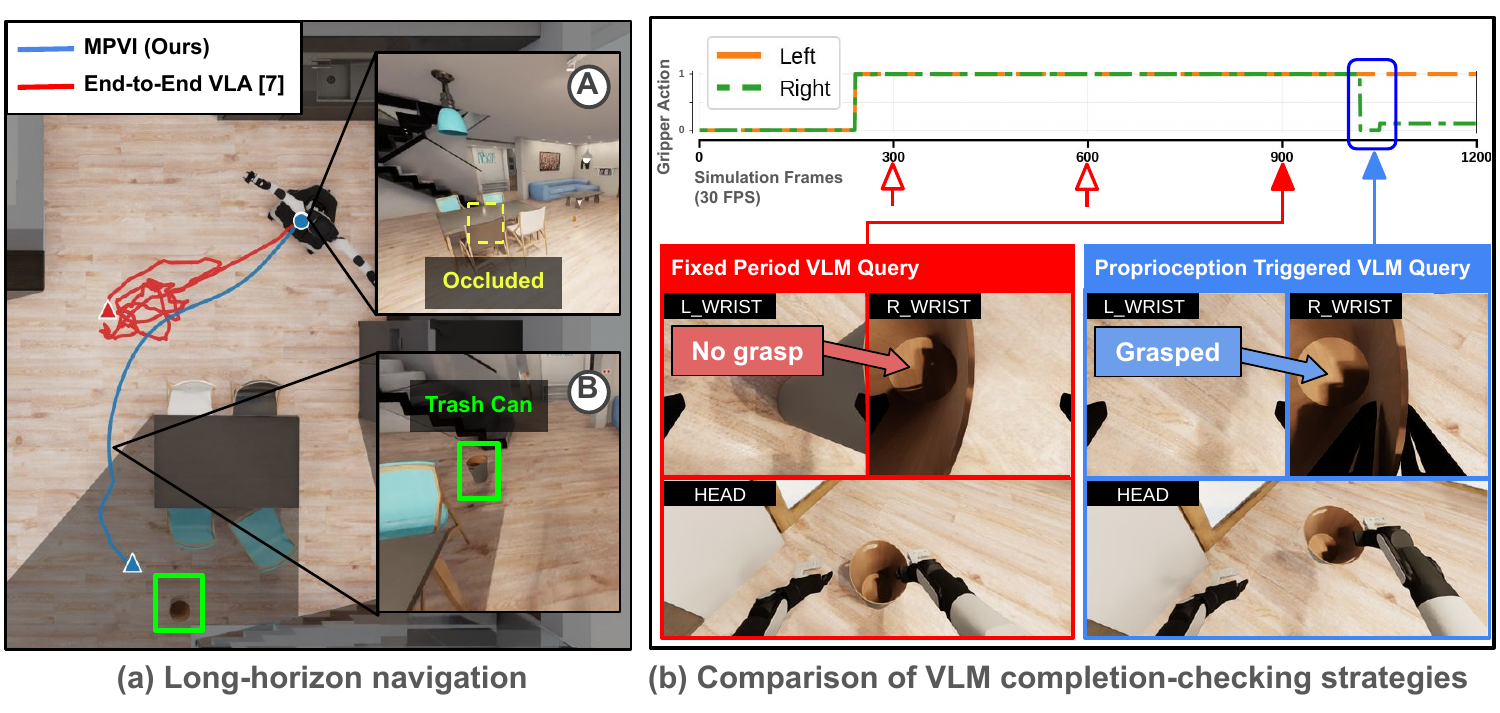}
    \caption{\textbf{(a)} Rollouts of the baseline VLA (red) and MPVI (blue) navigating toward a trash can initially occluded from the start pose (A). The
    baseline VLA fails to progress; MPVI combines open-vocabulary detection with frontier exploration (B) to locate and reach the target.
    \textbf{(b)} (Left, red) A fixed-period query at frame 900 passes all three camera views to the VLM. The trash
    can is visible in the head view, so the VLM returns a false positive judgment despite the gripper not having closed; this false positive cascades into subsequent subtasks. (Right, blue)
    Our proprioceptive trigger fires at frame 1020, once the right gripper closes and holds at a nonzero width, and directs the VLM to the corresponding wrist view (R\_WRIST), yielding
    reliable completion detection (gain quantified in \Cref{abl: proprio trigger}).}
    \label{fig: long-horizon navigation and proprio_vlm}
    \vspace{-1.5em} 
\end{figure*}

Reliable detection of manipulation subtask completion is essential for triggering handoff back to the motion planner.
Periodic VLM polling is unsuitable, as queries are computationally expensive and each query introduces an independent risk of hallucination, whereby a single false positive prematurely terminates manipulation and induces cascading failures in subsequent subtasks.
We therefore condition VLM queries on proprioceptive signals, triggering a completion check only when gripper or arm state indicates that a manipulation primitive such as a pick or place has just terminated.
Our implementation uses Gemini~2.5 Flash~\cite{gemini2.5} as the completion checker. \Cref{fig: long-horizon navigation and proprio_vlm}b illustrates the contrast between periodic polling and our proprioception-triggered query. Further design considerations are deferred to \Cref{app: completion checker details}.

\section{Experiments}
\label{sec: experiments}

% We now evaluate our method experimentally. 
In \Cref{exp: main} we study how \ourmethod compare to end-to-end VLA approaches on long-horizon mobile manipulation tasks from BEHAVIOR-1K.
Our hybrid approach increases task progress by 113\% over the end-to-end baseline.
We then consider how each component with the framework contributes to this improvement with an ablation study \Cref{exp: ablation}.
 % \begin{wrapfigure}{r}{0.63\textwidth}   % r=right, l=left, 45% column width
 %    \centering
 %    \vspace{-.2em}                       % optional: tighten top spacing
 %    \includegraphics[width=0.63\textwidth]{figures/VLM Query figure - sized - Google Slides.pdf}
 % \caption{Comparison of VLM completion-checking strategies on a ``pick up
 %  the trash can'' subtask. \textbf{(Left, red)} A fixed-period query at
 %  frame 900 passes all three camera views to the VLM, but does not indicate
 %  which wrist is performing the grasp. The can is visible in the head view,
 %  and the VLM returns a positive judgment despite the gripper not having
 %  closed on the object. This false positive cascades into subsequent
 %  subtasks. \textbf{(Right, blue)} Our proprioceptive trigger fires at frame
 %  1020, once the right gripper closes and holds at a nonzero width, and
 %  directs the VLM to attend to the corresponding wrist view (R\_WRIST), in
 %  which the grasped object is clearly visible. The proprioception-based
 %  querying strategy yields reliable completion detection, with the gain
 %  quantified in \Cref{abl: proprio trigger}.}
 %  \label{fig:proprio_vlm}

 %    \vspace{-1.2em}                       % optional: tighten bottom spacing
 %  \end{wrapfigure}

\subsection{Experiment Design}

\paragraph{Dataset and Baselines} 
We evaluate on the complete 50 task set of BEHAVIOR-1K~\cite{BEHAVIOR-1K} from the 2025 BEHAVIOR Challenge at NeurIPS~\cite{behavior2025challenge} in the OmniGibson simulator~\cite{igibson}, with 10 instances per task where the object positions and the initial robot pose are varied.
We compare against openpi-comet~\cite{bai2025comet}, the second-place solution at the challenge, which uses a whole-body VLA ($\pi_{0.5}$) and achieves an 11.40\% success rate; we adopt this baseline rather than the first-place solution~\cite{behavior-first-place} (12.40\% success rate) as the latter relies on human-engineered, task-specific heuristics that limit generalization beyond the evaluation tasks.
As comparing VLA performance requires carefully aligned testing conditions \cite{rasouli2026vlas} and openpi-comet reports best-of-$N$ results under specific evaluation settings, we rerun their released code within our own standardized framework to establish our baseline.
Furthermore, since our architecture incorporates their VLA as a module, our primary focus remains on the relative improvement achieved through our integration.

All experiments are run with a 24-core Intel Xeon Gold 6226 CPU, 128 GB RAM, and two NVIDIA Quadro RTX6000 GPUs (48 GB VRAM).

\subsection{\ourmethod vs. End-to-end VLA}
\label{exp: main}

% We compare our hybrid method, which integrates classical motion planning for navigation with a VLA for manipulation, against the end-to-end VLA baseline.
% We hypothesize that our method will outperform end-to-end learned VLAs in achieving task success on long-horizon mobile manipulation tasks.

\subsubsection{Metrics}

%% SUNDHAR'S ORIGINAL WRITING %%%%

% We adopt the four evaluation metrics introduced by the original BEHAVIOR-100 benchmark~\cite{behavior2021}:
% \begin{outline}[itemize]
%     \1 Q-Score: Fraction of goal BDDL predicates satisfied at episode end. Measures task success.
%     \1 Normalized simulation time: Total simulation time divided by average human teleoperation time for training instances of the same task.
%     \1 Base distance: Accumulated distance traveled by the robot base. Measures navigation efficiency.
%     \1 End-effector distance: Accumulated displacement of the robot's end effectors. Measures manipulation effort.
% \end{outline}

% Since raw distance and time metrics are confounded by task progress (a robot that completes more of the task will naturally travel farther and take longer), we additionally report efficiency metrics on episode pairs where both methods achieve matching Q-scores.
%%% END SUNDHAR'S ORIGINAL WRITING %%%%
We adopt the four metrics introduced by BEHAVIOR-100~\cite{behavior2021} and used for evaluation in the BEHAVIOR-1K challenge: Q-score (fraction of goal BDDL predicates satisfied at episode end), normalized simulation time (relative to human demonstration on the same task), base distance, and end-effector distance. Since raw distance and time are confounded by task progress, we additionally report these on episode pairs with matching Q-scores.

\subsubsection{Results and Discussion}
Complete results are presented in \Cref{appendix:full_results} and summarized in \Cref{fig:main_results}.
Gains concentrate on long-horizon tasks with many sub-goals across distinct locations, a regime in which Rasouli et al.~\cite{rasouli2026vlas} identify navigation failure, execution-order confusion, and abrupt termination as recurring failures modes.
Our method correspondingly does not improve performance on tasks where dexterous manipulation, like chopping vegetables, is the key bottleneck.
We attribute regression in task success to excessive motion planner intervention in scenarios where the VLA exhibits superior performance, such as maneuvering through narrow doorways while carrying objects.
Future work will address this limitation by developing adaptive intervention strategies triggered by deviations from the demonstration manifold and failure detection rather than relying on static subtask boundaries between navigation and manipulation.
\Cref{app: qualitative failures} details several failure modes we observed in the end-to-end VLA baseline that \ourmethod is able to address, adding a qualitative explanation to the quantitative results below.

\begin{wrapfigure}{r}{0.6\textwidth}  % “r” = right; 0.45\textwidth = width reserved for figure+caption
  \vspace{-20pt}                        % tweak vertical placement (optional)
  \centering
  \includegraphics[width=0.59\textwidth]{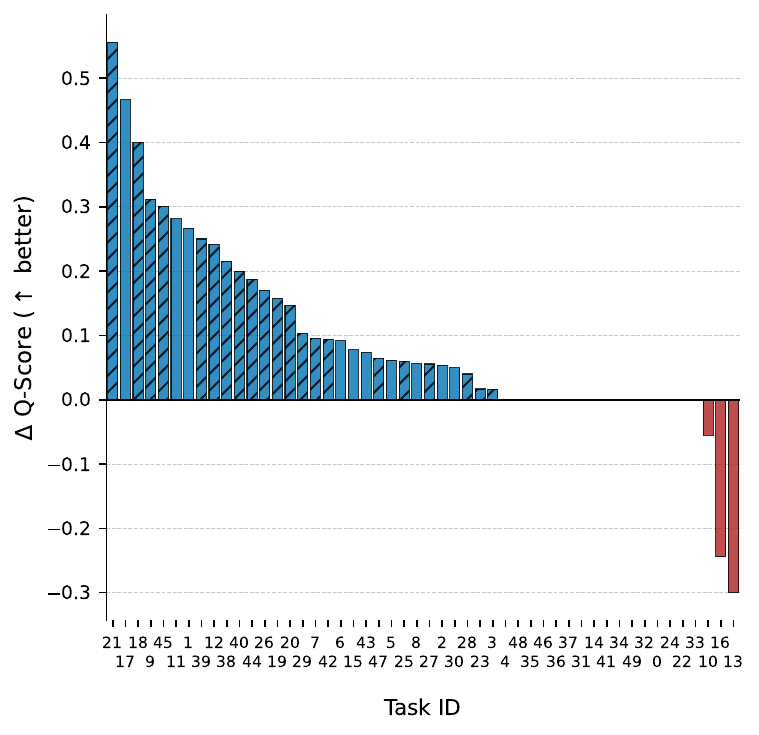}
  \caption{\textbf{Task progress improvement over baseline end-to-end VLA~\cite{bai2025comet}.} \ourmethod increases the mean Q-Score across all 50 tasks by 113\% with progress gains on 31 tasks (blue), matching the baseline on 16 tasks, and regressions on 3 tasks (red). Hatched bars indicate tasks in which the baseline did not complete any subtasks successfully (Q-score = 0).}
  \label{fig:main_results}
  \vspace{-10pt}                        % pull up the text a bit (optional)
\end{wrapfigure}

% Improvement in task success is most pronounced for tasks requiring navigation between distant objects, such as \texttt{picking\_up\_trash}.
% Tasks involving navigation through cluttered environments, such as \texttt{tidying\_bedroom} also show substantial improvement, as classical motion planning provides reliable navigation around obstacles in scenarios where the end-to-end VLA frequently fails to make progress.

% \input{tables/matched_q_metrics}

To isolate efficiency differences from the confounding effect of task progress, we compare episodes in which both methods achieved matching Q-scores ($N=20$ episodes).
Under this controlled comparison, our method exhibits 4.5\% lower base distance traveled and 9.8\% lower normalized simulation time.
These modest differences suggest that, when task progress is equivalent, the efficiency of navigation learned from human demonstrations is broadly comparable to that of motion-planner-based navigation.
Accordingly, the primary benefit of \ourmethod is not more efficient navigation in tasks where an end-to-end VLA would succeed, but improved robustness to failures that would otherwise degrade task performance.
While our method uses the VLA for manipulation, it achieves 13\% lower end-effector displacement, as it keeps the arm fixed during navigation, whereas the end-to-end VLA baseline produces unnecessary arm motion during navigation.

% \begin{figure*}[ht]
%     \centering
%     \includegraphics[width=\textwidth]{figures/qualitative_failure_analysis.pdf}
%     \caption{Three qualitative failure modes of the end-to-end VLA baseline that \ourmethod addresses: occlusion (left), premature termination (center), and execution-order confusion (right). \ourmethod resolves these via frontier exploration, navigation to the next manipulation subtask, and ordered subtask routing, respectively.}
%     \label{fig: qualitative failures}
% \end{figure*}

\subsection{Ablation Study}
\label{exp: ablation}

\begin{figure}[ht]
    \centering
    \includegraphics[width=\textwidth]{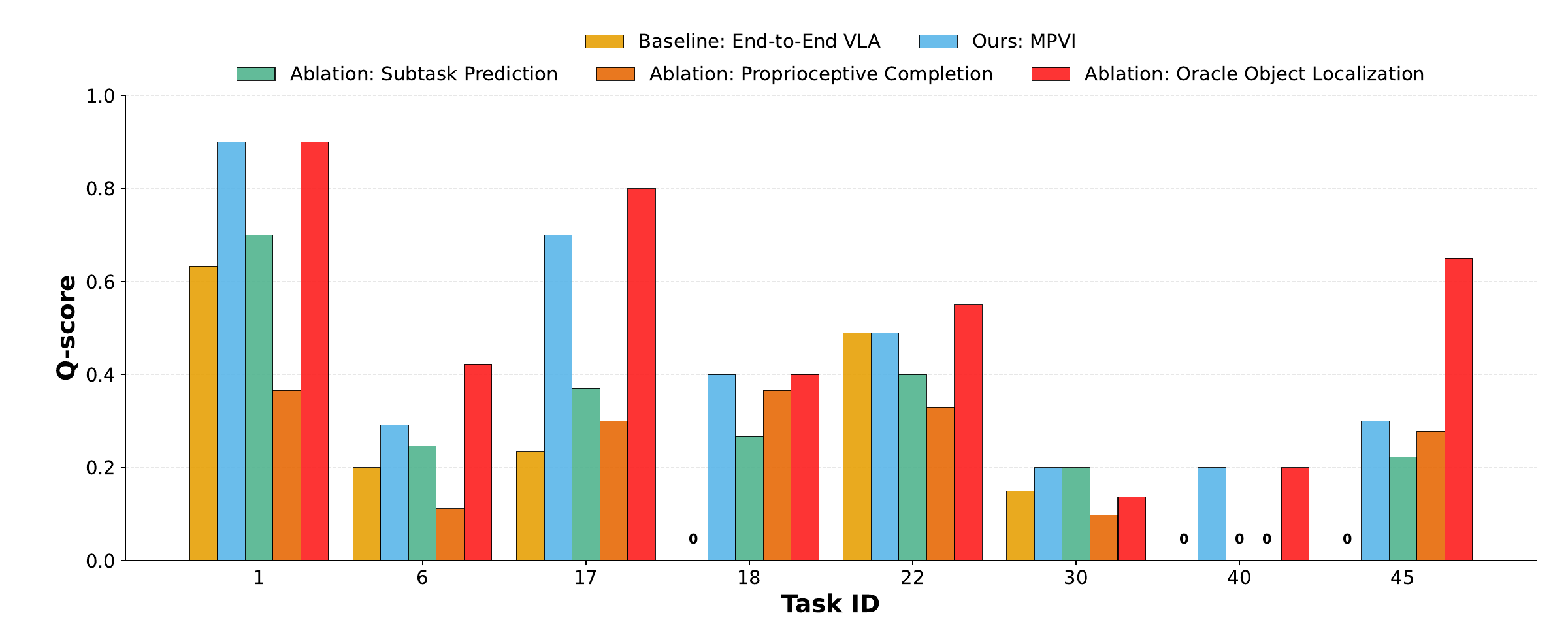}
    \caption{\textbf{Ablation Study Results.} Q-scores across a 12-task subset comparing our full method (MPVI) and the openpi-comet baseline to three system ablations.
    Tasks involving dexterous manipulation in which all methods achieve zero Q-score are omitted from the figure (\texttt{turning\_on\_radio, wash\_a\_baseball\_cap, hanging\_pictures, attach\_a\_camera\_to\_a\_tripod}). Relying only on \subtp{Subtask Prediction} drops performance by 31\% relative to MPVI, though it remains 41\% better than the baseline. Removing \compchk{Proprioceptive Completion} triggers decreases MPVI performance by 47\%. Isolating perception with \nav{Oracle Object Localization} improves Q-scores by 17\% over MPVI and 138\% over the baseline.
    }
    \label{fig:ablation}
    \vspace{-12pt}
\end{figure}

We conduct an ablation study to evaluate the contribution of three key components: classical motion planning for navigation beyond granular task decomposition alone (\Cref{sec: subtask,sec: integration}), our object detection and exploration procedure (\Cref{sec: motion planning}), and our proprioception-triggered completion checking (\Cref{sec: completion checker}).
\Cref{fig:ablation} reports their quantitative impact through resulting Q-scores for each variant.
We evaluate ablations on a subset of 12 tasks for which the baseline, openpi-comet \cite{bai2025comet}, provides a separate checkpoint finetuned only on demonstrations of these tasks rather than the complete task set.

\paragraph{\subtp{Subtask Prediction}}
To isolate the contribution of classical motion planning from that of granular task decomposition, we replace the motion planner with the VLA for all navigation subtasks, sequentially prompting it with subtask descriptions from our task planner and transitioning via our completion checker. This mirrors the ``subtask prediction'' approach in models like $\pi_{0.5}$~\cite{pi0.5}. Decomposition alone improves over the baseline but does not match our full method. Even with granular subtask descriptions, the VLA still fails at exploration and long-distance traversal through clutter which classical motion planning addresses.

%% SVS %%

% To isolate how much of our performance gain stems from classical motion planning versus granular task decomposition, we replace the motion planner with the VLA for all navigation subtasks and sequentially prompt it with natural language subtask descriptions from our task planner, transitioning via our completion checker.
% This mirrors the ``subtask prediction'' approach to long-horizon tasks in models like $\pi_{0.5}$ \cite{pi0.5}.
% While subtask prompting increases average Q-score by 73\% compared to the openpi-comet baseline, it decreased by 24\% relative to our full method.
% The VLA's continued struggle with exploration and long-distance traversal through clutter underscores the advantage of classical planning.

%% END SVS %%
\paragraph{\nav{Object Detection and Exploration}}
We replace our GroundingDINO-based object localization and search module with ground-truth object positions from the simulator. In most tasks, we observe minimal improvement in task success using ground truth object positions, suggesting that perception and exploration is not a significant bottleneck.
However, we observe a large improvement in Q-score with ground-truth positions in tasks taking place in a large, open, outdoor setting (e.g., \texttt{hiding\_Easter\_eggs}).
%% SVS %%

% We replace our GroundingDINO-based object localization and search procedure with ground truth object positions from the simulator.
% This Oracle ablation isolates the impact of perception and exploration failures on overall task success, resulting in a 129\% improvement in Q-score over the baseline and a 23\% increase over our method.
% In most tasks, we observe minimal improvement in task success using ground truth object positions, suggesting that perception and exploration is not a significant bottleneck.
% However, we observe a large improvement in Q-score with ground-truth positions in tasks taking place in a large, open, outdoor setting (e.g.,\texttt{hiding\_Easter\_eggs}).

%% END SVS %%
\paragraph{\compchk{Completion Checking}}
\label{abl: proprio trigger}
We replace proprioceptive-triggered VLM completion checking with periodic polling every 300 simulation steps (10\,s).
% We hypothesize that using proprioceptive signals like large changes in arm or gripper position to trigger VLM-based completion checking will improve detection of subtask completion, and thereby improve task success.
Removing the trigger produces two failure modes. False positives signal subtask completion prematurely, causing the robot to navigate to the next subtask location and leading to cascading failures.
False negatives prevent the framework from switching from the VLA back to the motion planner, effectively reducing our method to the end-to-end VLA since base motion is allowed during manipulation.
\section{Conclusion}
\label{sec: conclusion}
End-to-end Vision-Language-Action models underperform on long-horizon mobile manipulation despite training on large teleoperated expert demonstration datasets. We introduce \ourmethod, which interleaves a VLA with classical motion planning through LLM-based subtask decomposition, subtask orchestration, navigation grounded by open-vocabulary detection and frontier exploration, and proprioception-triggered VLM completion checking. Without additional data or training, \ourmethod improves mean Q-score by 113\% over a top-performing VLA baseline from the 2025 BEHAVIOR Challenge.

As Vision-Language-Action models continue to scale, we believe a key focus should be to develop principled criteria for when a learned policy should defer to a classical component, such as triggers based on failure detection or departure from the training distribution. Our results indicate that thoughtful integration of motion planning with a learned manipulation policy can address several recurring failure modes of end-to-end VLAs on long-horizon mobile manipulation. We hope this work opens new pathways for hybrid architectures, lowers the data and capacity demands placed on VLAs and brings long-horizon mobile manipulation closer to real-world deployment.

% We evaluate our method on the BEHAVIOR-1K dataset containing complex tasks in a realistic simulation environment.  
% We show that this integration yields significant improvement in task success and addresses end-to-end VLA failure modes.

% Future work will address several directions.
% Our method assumes access to a traversability map with labeled room boundaries.
% This assumption, while reasonable for many household settings, limits applicability to novel environments.
% Integrating vision-language navigation models~\citep{gu-etal-2022-vision} or constructing semantic maps online \cite{songsemantic} would relax this requirement and extend our approach to unmapped environments.

% Finally, while this work focuses on navigation, integration of classical methods for manipulation with learned policies may yield similar improvements within manipulation subtasks.
% More broadly, hybrid frameworks that leverage classical components may reduce the data and model capacity requirements of learned policies.

\section{Limitations}

\ourmethod has two main limitations. First, our framework assumes access to a traversability map with labeled room boundaries, which restricts deployment to environments mapped in advance. Online semantic mapping~\cite{songsemantic} or vision-language navigation~\cite{gu-etal-2022-vision} could lift this assumption. Second, our integration of classical planning is limited to navigation, with manipulation delegated entirely to the VLA. Performance therefore does not improve on tasks where the dominant bottleneck is dexterous manipulation. For example, neither the baseline nor \ourmethod achieves a nonzero Q-score on \texttt{attach\_a\_camera\_to\_a\_tripod} where the failure originates in the manipulation policy itself~\cite{rasouli2026vlas}. Extending the interleaving approach to manipulation, for example by routing pre-grasp motion to model-based motion planners, is a natural next step.

\clearpage
% The acknowledgments are automatically included only in the final and preprint versions of the paper.
% \acknowledgments{}

%===============================================================================

\bibliography{references}  % .bib

\newpage
\appendix
\crefalias{section}{appendix}
\crefalias{subsection}{appendix}
\crefalias{subsubsection}{appendix}
% \section{Appendix}

\section{Appendix: Full Results}
\label{appendix:full_results}

% \begin{figure*}[h]
%     \centering
%     \includegraphics[width=\textwidth]{figures/q_scores_density_columns.pdf}
%     \caption{Ablation study comparing task success (Q-Score) across eight BEHAVIOR-1K tasks. Bars represent Q-scores achieved by each task instance, with black horizontal lines indicating the mean. Replacing our object detection with ground truth object localization (Oracle Localization) yields modest gains, indicating perception is not a primary bottleneck. Removing proprioceptive-triggered completion checking in favor of periodic VLM queries (Ours w/o Proprio. Trigger) degrades performance across most tasks. Tasks where all methods achieve 0 Q-Score are omitted.
% }
%     \label{fig:ablation_study}
% \end{figure*}

\begin{table}[htbp]
\centering
\small
\caption{Q-Score Per Task}
\label{tab:per_task_q_score}
\begin{tabular}{llccc}
\toprule
Task ID & Task & End-to-End VLA & Ours & $\Delta$ Q \\
\midrule
21 & collecting\_childrens\_toys & 0.00 & 0.56 & 0.56 \\
17 & bringing\_water & 0.23 & 0.70 & 0.47 \\
18 & tidying\_bedroom & 0.00 & 0.40 & 0.40 \\
9 & putting\_up\_Christmas\_decorations\_inside & 0.00 & 0.31 & 0.31 \\
45 & cook\_hot\_dogs & 0.00 & 0.30 & 0.30 \\
11 & putting\_dishes\_away\_after\_cleaning & 0.11 & 0.39 & 0.28 \\
1 & picking\_up\_trash & 0.63 & 0.90 & 0.27 \\
39 & spraying\_fruit\_trees & 0.00 & 0.25 & 0.25 \\
12 & preparing\_lunch\_box & 0.00 & 0.24 & 0.24 \\
38 & spraying\_for\_bugs & 0.00 & 0.21 & 0.21 \\
40 & make\_microwave\_popcorn & 0.00 & 0.20 & 0.20 \\
44 & chopping\_wood & 0.00 & 0.19 & 0.19 \\
26 & assembling\_gift\_baskets & 0.00 & 0.17 & 0.17 \\
19 & outfit\_a\_basic\_toolbox & 0.00 & 0.16 & 0.16 \\
20 & sorting\_vegetables & 0.00 & 0.15 & 0.15 \\
29 & clean\_up\_your\_desk & 0.00 & 0.10 & 0.10 \\
7 & picking\_up\_toys & 0.00 & 0.10 & 0.10 \\
42 & chop\_an\_onion & 0.00 & 0.09 & 0.09 \\
6 & hiding\_Easter\_eggs & 0.20 & 0.29 & 0.09 \\
15 & bringing\_in\_wood & 0.03 & 0.11 & 0.08 \\
43 & slicing\_vegetables & 0.00 & 0.07 & 0.07 \\
47 & freeze\_pies & 0.00 & 0.06 & 0.06 \\
5 & setting\_mousetraps & 0.38 & 0.44 & 0.06 \\
25 & clearing\_food\_from\_table\_into\_fridge & 0.00 & 0.06 & 0.06 \\
8 & rearranging\_kitchen\_furniture & 0.10 & 0.16 & 0.06 \\
27 & sorting\_household\_items & 0.00 & 0.06 & 0.06 \\
2 & putting\_away\_Halloween\_decorations & 0.32 & 0.37 & 0.05 \\
30 & setting\_the\_fire & 0.15 & 0.20 & 0.05 \\
28 & getting\_organized\_for\_work & 0.00 & 0.04 & 0.04 \\
23 & boxing\_books\_up\_for\_storage & 0.00 & 0.02 & 0.02 \\
3 & cleaning\_up\_plates\_and\_food & 0.00 & 0.02 & 0.02 \\
48 & canning\_food & 0.00 & 0.00 & 0.00 \\
4 & can\_meat & 0.00 & 0.00 & 0.00 \\
32 & wash\_a\_baseball\_cap & 0.00 & 0.00 & 0.00 \\
37 & clean\_a\_trumpet & 0.00 & 0.00 & 0.00 \\
31 & clean\_boxing\_gloves & 0.00 & 0.00 & 0.00 \\
36 & clean\_a\_patio & 0.00 & 0.00 & 0.00 \\
35 & attach\_a\_camera\_to\_a\_tripod & 0.00 & 0.00 & 0.00 \\
41 & cook\_cabbage & 0.00 & 0.00 & 0.00 \\
46 & cook\_bacon & 0.00 & 0.00 & 0.00 \\
49 & make\_pizza & 0.00 & 0.00 & 0.00 \\
33 & wash\_dog\_toys & 0.00 & 0.00 & 0.00 \\
0 & turning\_on\_radio & 0.00 & 0.00 & 0.00 \\
34 & hanging\_pictures & 0.00 & 0.00 & 0.00 \\
22 & putting\_shoes\_on\_rack & 0.49 & 0.49 & 0.00 \\
24 & storing\_food & 0.00 & 0.00 & 0.00 \\
14 & carrying\_in\_groceries & 0.00 & 0.00 & 0.00 \\
10 & set\_up\_a\_coffee\_station\_in\_your\_kitchen & 0.20 & 0.15 & -0.06 \\
16 & moving\_boxes\_to\_storage & 0.80 & 0.56 & -0.24 \\
13 & loading\_the\_car & 0.30 & 0.00 & -0.30 \\
\bottomrule
\end{tabular}
\end{table}
\begin{table}[htbp]
\centering
\setlength{\tabcolsep}{3pt}
\caption{Ablation Study}
\label{tab:q_score_ablation}
\begin{tabular}{lcccccc}
\toprule
Task ID & Task & \multicolumn{5}{c}{Q-Score} \\
\cmidrule(lr){3-7}
 &  & E2E VLA & Ours & Oracle Loc. & Subtask Pred.& No Prop. Trigger \\
\midrule
0 & turning\_on\_radio & 0.00 & 0.00 & 0.00 & 0.00 & 0.00 \\
1 & picking\_up\_trash & 0.63 & 0.90 & 0.90 & 0.70 & 0.37 \\
6 & hiding\_Easter\_eggs & 0.20 & 0.29 & 0.42 & 0.25 & 0.11 \\
17 & bringing\_water & 0.23 & 0.70 & 0.80 & 0.37 & 0.30 \\
18 & tidying\_bedroom & 0.00 & 0.40 & 0.40 & 0.27 & 0.37 \\
22 & putting\_shoes\_on\_rack & 0.49 & 0.49 & 0.55 & 0.40 & 0.33 \\
30 & setting\_the\_fire & 0.15 & 0.20 & 0.14 & 0.20 & 0.10 \\
32 & wash\_a\_baseball\_cap & 0.00 & 0.00 & 0.00 & 0.00 & 0.00 \\
34 & hanging\_pictures & 0.00 & 0.00 & 0.00 & 0.00 & 0.00 \\
35 & attach\_a\_camera\_to\_a\_tripod & 0.00 & 0.00 & 0.00 & 0.00 & 0.00 \\
40 & make\_microwave\_popcorn & 0.00 & 0.20 & 0.20 & 0.00 & 0.00 \\
45 & cook\_hot\_dogs & 0.00 & 0.30 & 0.65 & 0.22 & 0.28 \\
\bottomrule
\end{tabular}
\end{table}
% \input{tables/dist_base}
% \newpage
% \input{tables/dist_ee}
% \input{tables/norm_time}

\newpage
\section{Appendix: Qualitative Failure Analysis}
\label{app: qualitative failures}

\begin{figure*}[h]
    \centering
    \includegraphics[width=\textwidth]{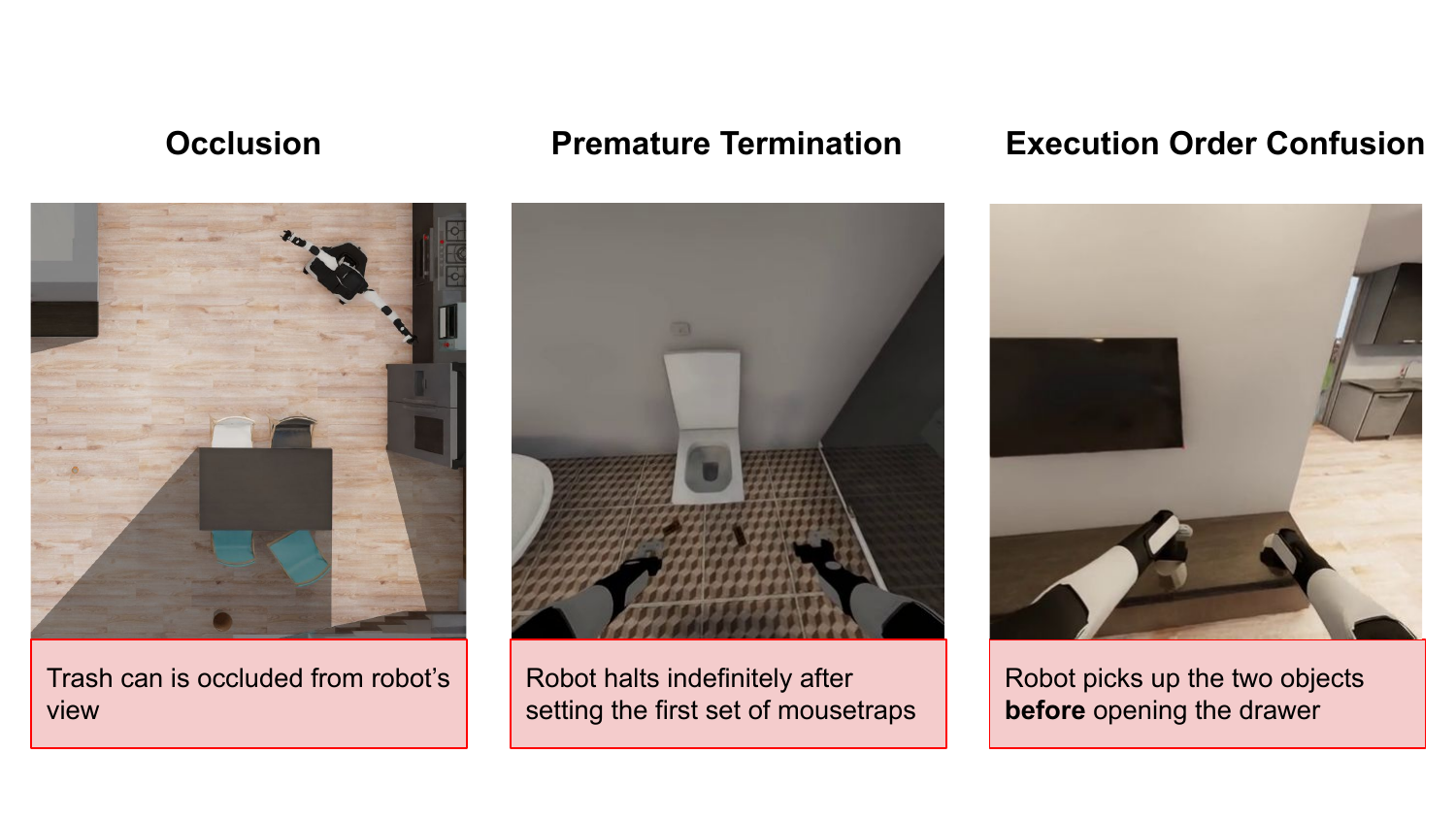}
    \caption{Three qualitative failure modes of the end-to-end VLA baseline that \ourmethod addresses: occlusion (left), premature termination (center), and execution-order confusion (right). \ourmethod resolves these via frontier exploration, navigation to the next manipulation subtask, and ordered subtask routing, respectively.}
    \label{fig: qualitative failures}
\end{figure*}

Rasouli et al.~\cite{rasouli2026vlas} identify navigation failures, abrupt termination, and execution-order confusion as recurring failure modes of end-to-end VLAs on long-horizon mobile manipulation. \Cref{fig: qualitative failures} shows one openpi-comet rollout per failure mode.

\paragraph{Occlusion.} The left panel shows a rollout of \texttt{picking\_up\_trash}. The trash can, the first manipulation target, is occluded from the robot's initial viewpoint. The end-to-end VLA does eventually locate the trash can, but only after extended wandering that consumes the episode budget before any subsequent subtask can be attempted. \ourmethod replaces this with open-vocabulary object detection, falling back to frontier exploration if detection fails. The trash can is localized in time to complete the downstream collect-and-place subtasks.

\paragraph{Premature termination.} The center panel shows \texttt{setting\_the\_mousetrap}. The task requires placing one set of traps under the toilet, then retrieving two more from a cabinet and placing them under the sink. 
After placing the first traps under the toilet, the end-to-end VLA abruptly terminates and makes no further progress toward task success for the remainder of the episode, despite being far from the time limit. \ourmethod avoids this outcome by returning control to the motion planner once a manipulation segment completes, which then steers the robot to the cabinet where the VLA then can be invoked again to continue the progress.

\paragraph{Execution-order confusion.} The right panel shows \texttt{putting\_away\_Halloween\_decoration}. The correct order is (i) open the cabinet, (ii) pick up two candles, and (iii) place the candles in the cabinet. The end-to-end VLA picks up both candles before opening the cabinet. With both grippers occupied, the robot has no free hand to operate the cabinet door, and no further task progress occurs within the episode. \ourmethod prevents this through its subtask planner, which decomposes the instruction $\instruction$ into an ordered sequence of short subtasks, and its orchestrator, which dispatches each subtask in order and verifies completion before advancing

%%%%%%%%%%%%%%%%%%%%%%%%%%%%%%%%%%%%%%%%%%%%%%%%%%%%%%%%%%%%%%%%%%%%%%%%%%%%%%%
%%%%%%%%%%%%%%%%%%%%%%%%%%%%%%%%%%%%%%%%%%%%%%%%%%%%%%%%%%%%%%%%%%%%%%%%%%%%%%%
\section{Appendix: Method Implementation Details}
\label{app: method details}

This appendix supplements the details of how each of the fours modules were implemented from \Cref{sec: method}.

\subsection{Subtask Planning}
\label{app: subtask planning details}

We prompt an LLM with the natural-language task description $\instruction$ to segment a long-horizon task into short-horizon subtasks; the full prompt is in \Cref{appendix:prompts}.
To improve plan quality, we fine-tune the LLM on human-annotated task plans from the BEHAVIOR-1K training dataset.
The LLM outputs a plan $\plan = \{\segment_1, \ldots, \segment_n\}$ where each segment $\segment_i = (\category_i, \subtaskdescription_i, \target_i, \completioncriteria_i)$ consists of a category $\category_i \in \{\textsc{navigation}, \textsc{manipulation}\}$, a natural-language subtask description $\subtaskdescription_i$, a target $\target_i$ for navigation segments specifying where to navigate, and a completion criterion $\completioncriteria_i$.
The completion criterion describes the desired world state (e.g., ``there are two soda cans in the trash can'') rather than a desired action (e.g., ``the robot put the soda can in the trash can'').

A post-processing step extracts additional metadata used for subtask completion detection.
First, we check whether each manipulation subtask can be mapped to a manipulation primitive (e.g., pick or place) with a pre-defined completion heuristic.
Second, we label manipulation subtasks that immediately precede navigation subtasks, since detecting completion of these subtasks is necessary to trigger the transition to the motion planner.
This information is consumed by the completion checker (\Cref{app: completion checker details}).

\paragraph{Relation to prior work.}
End-to-end VLA methods, including $\pi_{0.5}$, incorporate built-in subtask prediction capabilities and provide these predictions as natural-language inputs to the policy~\cite{pi0.5}.
Symbolic planning methods instead generate sequences of predefined affordances or skills~\cite{ahn2022can, singh_progprompt_2022}.
In contrast, we leverage the task plan to coordinate execution between the VLA and the motion planner.
Utilizing $\pi_{0.5}$'s built-in subtask prediction could potentially replace our LLM-based planning; however, this capability was not included in the open-source code release, so we leave this investigation to future work.

\subsection{Orchestrator Details}
\label{app: orchestrator details}

Given the plan $\plan$, the orchestrator groups segments into execution units: navigation segments are executed individually, while consecutive manipulation segments are grouped into a single execution unit.
For each navigation segment $\segment_i$ the orchestrator invokes the navigation module with target $\target_i$.
For each manipulation group, the orchestrator invokes the manipulation module with the concatenated subtask descriptions $\{\subtaskdescription_j, \ldots, \subtaskdescription_k\}$ and the completion criterion $\completioncriteria_k$ of the final segment in the group, where $j$ and $k$ are the indices of the first and last manipulation segments in the group, respectively.

The orchestrator maintains a knowledge base that stores the locations of rooms, areas, and large fixtures, as well as object locations discovered during execution.
When the navigation module successfully localizes a target object, its position is added to the knowledge base, enabling subsequent navigation subtasks targeting the same object to plan paths directly without repeating the search procedure.
The orchestrator additionally tracks subtask completion status; if a module fails to complete its subtask, the orchestrator marks the subtask as failed and proceeds to the next segment in the plan.

\subsection{Navigation}
\label{app: navigation details}

Navigation subtasks are executed using classical motion planning rather than the VLA.
Specifically, the navigation module consists of an A* path planner and a pure-pursuit controller which control base movement.
The robot's arms are maintained in a fixed configuration during navigation.

A fundamental challenge in employing classical motion planning is the requirement for information not available to end-to-end VLA approaches: a traversability map and target object locations.
We assume access to a traversability map with labeled rooms, and accurate robot localization.
These assumptions are reasonable for household environments, where robust semantic mapping and localization methods are well-established~\cite{raychaudhuri2025semantic, songsemantic}.

\paragraph{Target localization.}
Target object localization presents a greater challenge as object positions are not known a priori.
We first determine whether the robot is already within the handoff distance $d_h$ of the target object $\target$; if so, navigation is complete.
Otherwise, we query a knowledge base maintained by the system using $\target$ as a key.
The knowledge base stores the locations of rooms and large fixtures such as refrigerators and countertops.
If the location of $\target$ is known, a path is planned directly to it.

When the location of $\target$ is unknown, the robot navigates to the room or area specified in $\subtaskdescription$ and attempts to localize the object visually.
We employ GroundingDINO~\cite{liu2024grounding} with $\target$ as the text query to detect the target object in the robot's head camera image.
Upon successful detection, the bounding-box center is transformed from the camera frame to the world frame using the camera pose and intrinsics.
The depth of the object is obtained by computing the median depth within the bounding box from the corresponding depth image.
If the object is not detected, the robot performs frontier-based exploration.
The planner maintains a visibility map of observed and unobserved free space by projecting the robot's field of view onto the occupancy map using ray-based visibility updates, and selects actions that increase coverage of previously unseen regions.
At each iteration, the robot first reorients toward unexplored space visible from its current position, then plans a path using A* to the nearest reachable frontier between observed and unobserved space.
Visual detection of all task-relevant objects is performed continuously throughout exploration, and newly detected objects are opportunistically localized and cached for future subtasks.

Navigation terminates when the robot is within the handoff distance $d_h$ of $\target$, at which point control transitions to the VLA for manipulation.
Robot base repositioning within the handoff distance, which is not a task or object specific parameter, is handled by the VLA.

\subsection{Completion Checker}
\label{app: completion checker details}

A VLM-based completion checker receives the head-camera image and the completion criterion $\completioncriteria$ and returns a binary completion judgment; the completion-checking prompt is in \Cref{appendix:prompts}.
VLM hallucinations present a significant risk: false positives terminate manipulation prematurely, potentially causing cascading failures in subsequent subtasks.
Three design choices mitigate this risk.

\paragraph{State-based completion criteria.}
Completion criteria are formulated as state-based rather than action-based predicates.
For example, verifying whether ``there is one soda can in the trash can'' from a single image is more reliable than verifying whether ``the robot placed the soda can in the trash can,'' which requires temporal reasoning about actions that may not be observable in a static frame or sequence of frames.

\paragraph{Subtask sequencing.}
While the planner generates granular manipulation subtasks, completion is verified only for the terminal state preceding a navigation subtask.
For instance, a plan may contain sequential subtasks ``pick up the soda can'' followed by ``place the soda can in the trash can,'' but the VLM is queried only to verify that ``there is one soda can in the trash can'' prior to transitioning to navigation.
This reduces the frequency of VLM queries and avoids verification of intermediate states that may be difficult to assess visually.

\paragraph{Proprioceptive trigger.}
Proprioceptive signals, rather than periodic polling, trigger VLM completion checks.
A set of general proprioceptive heuristics monitor gripper and arm state for all manipulation subtasks, with additional heuristics applied when a manipulation primitive has been identified during post-processing.
For example, a pick primitive initiates a completion check when gripper and arm state indicate a successful grasp, while a place primitive initiates a check upon object release.
This ensures that VLM queries occur at semantically meaningful moments and reduces unnecessary calls.
If the VLM does not confirm completion, the VLA continues execution until a timeout is reached.

\section{Appendix: VLM Prompts}
\label{appendix:prompts}

\paragraph{Task Planning Prompt.}
\begin{verbatim}
You are creating a task plan for a robot based on a task description.

Task description: {task_description}

The robot is a bimanual mobile robot that can:
- Navigate to different areas/rooms in a house
- Navigate to specific objects
- Manipulate objects using two arms with grippers

Create a plan consisting of navigation and manipulation segments.

For each segment, determine:
1. Type: "navigation" (move to a location) or "manipulation" (interact with objects)
2. Description: Brief description of the action
3. Completion criteria: Observable condition indicating segment is complete
(describe the END STATE)
4. Target: For navigation segments, the destination

IMPORTANT for completion criteria:
- Describe the terminal state, NOT the action
- Good: "The object is in the container"
- Bad: "The robot puts the object in the container"

Output as JSON:
```json
{
  "task_name": "short task name",
  "segments": [
    {
      "type": "navigation",
      "description": "Navigate to the kitchen",
      "completion_criteria": "The robot is in the kitchen area",
      "target": "kitchen"
    },
    {
      "type": "manipulation",
      "description": "Pick up the object",
      "completion_criteria": "The object is held in the robot's gripper",
      "target": null
    }
  ]
}
```
\end{verbatim}

\paragraph{VLM Completion Checking Prompt.}
\begin{verbatim}
You are checking if a robot manipulation task is complete.

Task: {description}
Completion criteria: {completion_criteria}

Look at the image and determine if the completion criteria has been met.

Answer with:
- "YES, the task is complete" if the criteria is clearly satisfied
- "NO, the task is not complete" if the criteria is not yet satisfied

Be strict: only say YES if you are confident the criteria is met.

Your answer:
\end{verbatim}

\end{document}